\documentclass[sigconf]{acmart}

\usepackage{booktabs} 
\usepackage{subfigure}
\usepackage{epstopdf}
\usepackage{enumitem}
\usepackage[algo2e,ruled,vlined,linesnumbered]{algorithm2e}
\usepackage{multirow}
\usepackage{xcolor,colortbl}

\newcommand{\dosemic}{\renewcommand{\@endalgocfline}{\algocf@endline}}
\let\oldnl\nl
\newcommand{\nonl}{\renewcommand{\nl}{\let\nl\oldnl}}

\begin{document}

\copyrightyear{2017}
\acmYear{2017}
\setcopyright{acmcopyright}
\acmConference{KDD'17}{}{August 13--17, 2017, Halifax, NS, Canada.} \acmPrice{15.00}
\acmDOI{http://dx.doi.org/10.1145/3097983.3098115}
\acmISBN{978-1-4503-4887-4/17/08}

\fancyhead{}
\settopmatter{printacmref=false, printfolios=false}
	
\title{A Context-aware Attention Network for Interactive \\ Question Answering}

\author{Huayu Li$^{1}$, Martin Renqiang Min$^2$, Yong Ge$^3$, Asim Kadav$^2$}
\titlenote{Most of this work was done when the first author was an intern at NEC Labs America.}
\affiliation{\institution{$^1$Department of Computer Science, UNC Charlotte}}
\affiliation{\institution{$^2$Machine Learning Group, NEC Laboratories America}}
\affiliation{\institution{$^3$Management Information Systems, University of Arizona}}
\email{hli38@uncc.edu, {renqiang,asim}@nec-labs.com, yongge@email.arizona.edu.}

\renewcommand{\shortauthors}{}
\renewcommand{\shorttitle}{}

\begin{abstract}
Neural network based sequence-to-sequence models in an encoder-decoder framework have been successfully applied to solve Question Answering (QA) problems, predicting answers from statements and questions. However, almost all previous models have failed to consider detailed context information and unknown states under which systems  do not have enough information to answer given questions. These scenarios with incomplete or ambiguous information are very common in the setting of Interactive Question Answering (IQA). To address this challenge, we develop a novel model, employing context-dependent word-level attention for more accurate statement representations and question-guided sentence-level attention for better context modeling. We also generate unique IQA datasets to test our model, which will be made publicly available. Employing these attention mechanisms, our model accurately understands when it can output an answer or when it requires generating a supplementary question for additional input depending on different contexts. When available, user's feedback is encoded and directly applied to update sentence-level attention to infer an answer. Extensive experiments on QA and IQA datasets quantitatively demonstrate the effectiveness of our model with significant improvement over state-of-the-art conventional QA models.
\end{abstract}

%
%


\maketitle

\section{Introduction}

With the availability of large-scale QA datasets, high-capacity machine learning/data mining models, and powerful computational devices, research on QA has become active and fruitful. Commercial QA products such as Google Assistant, Apple Siri, Amazon Alexa, Facebook M, Microsoft Cortana, Xiaobing in Chinese, Rinna in Japanese, and MedWhat have been released in the past several years. The ultimate goal of QA research is to build intelligent systems capable of naturally communicating with humans, which poses a major challenge for natural language processing and machine learning. Inspired by recent success of sequence-to-sequence models with an encoder-decoder framework \citep{Ilya:2014:NIPS, Kyunghyun:2014:EMNLP}, researchers have attempted to apply variants of such models with explicit memory and attention to QA tasks, aiming to move a step further from machine learning to machine reasoning~\citep{Sukhbaatar:2015:NIPS,Ankit:2016:ICML,Caiming:2016:ICML}. Similarly, all these models employ encoders to map statements and questions to fixed-length feature vectors, and a decoder to generate outputs.  Empowered by the adoption of memory and attention, they have achieved remarkable success on several challenging public datasets, including the recently acclaimed Facebook bAbI dataset~\cite{Jason:2015:CoRR}.

However, previous models suffer from the following important limitations~\cite{Caiming:2016:ICML,Ankit:2016:ICML,Sukhbaatar:2015:NIPS,Jason:2014:CoRR}. First, they fail to model context-dependent meaning of words. Different words may have different meanings in different contexts, which increases the difficulty of extracting the essential semantic logic flow of each sentence in different paragraphs. Second, many existing models only work in ideal QA settings and fail to address the uncertain situations under which models require additional user input to gather complete information to answer a given question. As shown in Table~\ref{table:QA_diff}, the example on the top is an ideal QA problem. We can clearly understand what the question is and then locate the relevant input sentences to generate the answer. But it is hard to answer the question in the bottom example, because there are two types of bedrooms mentioned in all input sentences (i.e., the story) and we do not know which bedroom the user refers to. These scenarios with incomplete information naturally appear in human conversations, and thus, effectively handling them is a key capability of intelligent QA models.

\begin{figure}[t]
	\centering
	\includegraphics[width=0.48\textwidth]{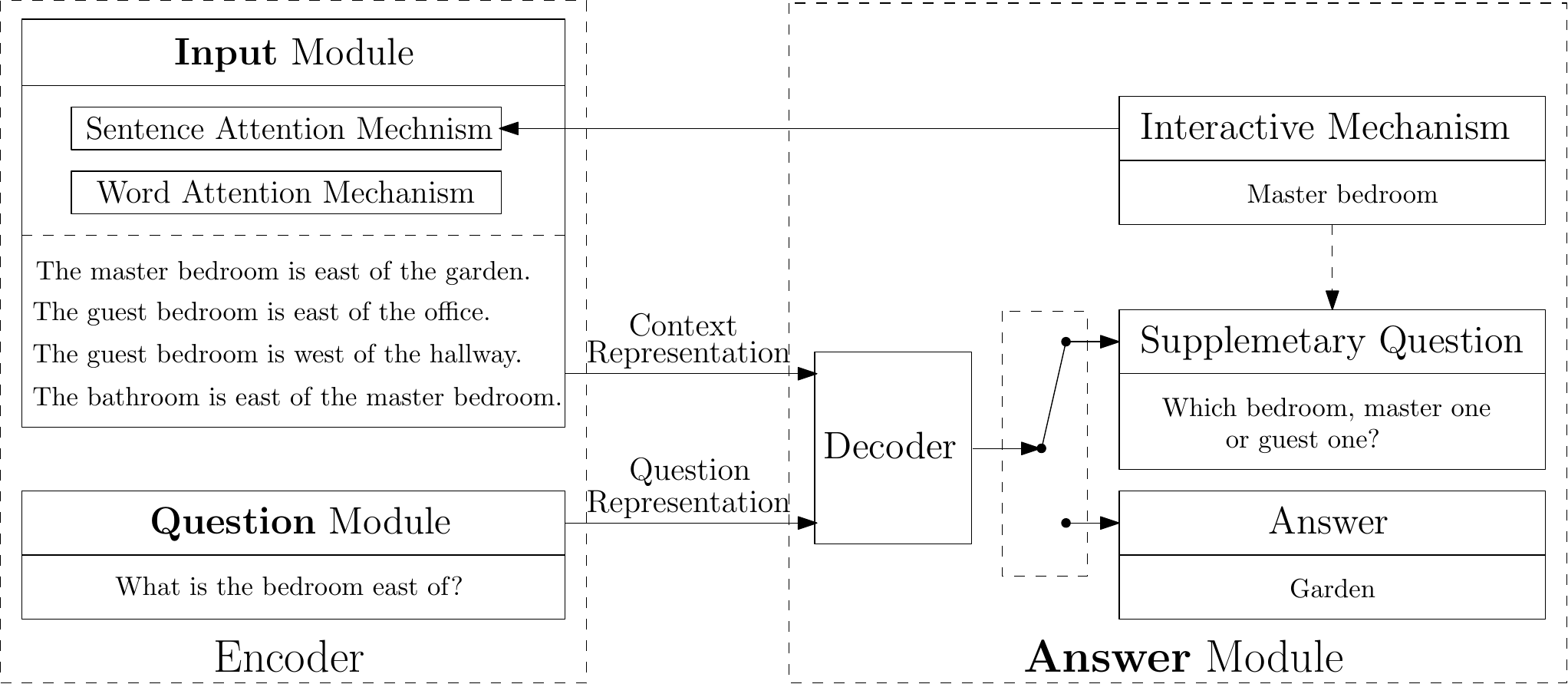}
	\vspace{-0.7cm}
	\caption{An example of QA problem using CAN.}
	\label{Fig:Exm_Frame}
	\vspace{-0.7cm}
\end{figure}

To address the challenges presented above, we propose a Context-aware Attention Network (CAN) to learn fine-grained representations for input sentences, and develop a mechanism to interact with user to comprehensively understand a given question. Specifically, we employ two-level attention applied at word level and sentence level to compute representations of all input sentences. The context information extracted from an input story is allowed to influence the attention over each word, and governs the word semantic meaning contributing to a sentence representation. In addition, an interactive mechanism is created to generate a supplementary question for the user when the model feels that it does not have enough information to answer a given question. User's feedback for the supplementary question is then encoded and exploited to attend over all input sentences to infer an answer. Our proposed model CAN can be viewed as an encoder-decoder approach augmented with two-level attention and an interactive mechanism, rendering our model self-adaptive, as illustrated in Figure~\ref{Fig:Exm_Frame}.

Our contributions in this paper are summarized as follows:
\begin{itemize}[leftmargin=*,itemsep=-1pt]
\item We develop a new encoder-decoder model called CAN for QA with two-level attention. Owing to the new attention mechanism, our model avoids the necessity of tuning-sensitive multiple-hop attention that is required by previous QA models such as MemN2N~\cite{Sukhbaatar:2015:NIPS} and DMN+~\cite{Caiming:2016:ICML}, and knows when it can readily output an answer and when it needs additional information from user depending on different contexts.
\item We augment the encoder-decoder framework for QA with an interactive mechanism for handling user's feedback, which immediately changes sentence-level attention to infer a final answer without additional model training.  To the best of our knowledge, our work is the first to augment the encoder-decoder framework to explicitly model unknown states with incomplete or ambiguous information for IQA and the first to propose the IQA concept to improve QA accuracy.
\item  We generate a new dataset based on the Facebook bAbI dataset, namely ibAbI, covering several representative IQA tasks. We make this dataset publicly available to the community, which could provide a useful resource for others to continue studying IQA problems.
\item  We conduct extensive experiments to show that our approach outperforms state-of-the-art models on both QA and IQA datasets. Specifically, our approach achieves $40\%$ improvement over conventional QA models without an interactive procedure (e.g., MemN2N and DMN+) on IQA datasets.
\end{itemize}

\begin{table} [t]
	\centering
	\begin{tabular}{|l|}
		\hline
		The office is north of the kitchen. \\
		The garden is south of the kitchen. \\
		Q: What is north of the kitchen? \\
		A: Office \\
		\hline
		\hline
		The master bedroom is east of the garden. \\
		The guest bedroom is east of the office. \\
		Q: What is the bedroom east of? \\
		A: Unknown \\
		\hline
	\end{tabular}
	\caption{Two examples of QA problem (there are two input sentences before each question). Top is an ideal QA example, where question is very clear. Bottom is an example with incomplete information, where question is ambiguous and it is difficult to provide an answer only using input sentences.}
	\vspace{-0.8cm}
	\label{table:QA_diff}
\end{table}

\vspace{-0.3cm}
\section{Related Work}
Recent work on QA has been heavily influenced by research on various neural network models with attention and/or memory in an encoder-decoder framework. These models have been successfully applied to image classification~\citep{Paul:2016:CoRR}, image captioning~\citep{Volodymyr:2014:NIPS}, machine translation~\citep{Kyunghyun:2014:EMNLP, Dzmitry:2015:ICLR,Minh:2015:CoRR}, document classification~\citep{Zichao:2016:HLT}, and textual/visual QA~\citep{Sukhbaatar:2015:NIPS,Zichao:2015:CoRR,Jiasen:2016:CoRR,Ankit:2016:ICML,Caiming:2016:ICML}.  For textual QA in the form of statements-question-answer triplets, MemN2N~\cite{Sukhbaatar:2015:NIPS} maps each input sentence to an input representation space regarded as a memory component. The output representation is calculated by summarizing over input representations with different attention weights. This single-layer memory is extended to multi-layer memory by reasoning the statements and the question with multiple hops. Instead of simply stacking the memory layers, Dynamic Memory Network (DMN) updates memory vectors through a modified GRU~\cite{Ankit:2016:ICML}, in which the gate weight is trained in a supervised fashion. To improve DMN by training without supervision, DMN+~\cite{Caiming:2016:ICML} encodes input sentences with a bidirectional GRU and then utilizes an attention-based GRU to summarize these input sentences. Neural Turing Machine (NTM)~\citep{Alex:2014:CoRR}, a model with content and location-based memory addressing mechanisms, has also been used for QA tasks recently. There is other recent work about QA using external resources~\citep{Anthony:2014:KDD, Denis:2016:SIGIR, Karl:2015:NIPS, David:2016:CoRR,PengchengYin:2015:CIKM,DenisSavenkov:2016:SIGIR}, and exploring dialog tasks~\citep{Jason:2016：NIPS, Jason:2016:CoRR_2,OriolVinyals:2015:CoRR}. Both MemN2N and DMN+ do not model context-aware word attention, instead, they use multi-hop memory. However, the QA performance produced by MemN2N and DMN+ is very sensitive to the number of hops.

In contrast, our proposed model is context-aware and self-adaptive. It avoids multiple-hop attention and knows when to output an answer and when to request additional information from a user. In addition, our IQA model works on conventional textual statement-question-answer triplets and effectively solves conventional QA problems with incomplete or ambiguous information. These IQA tasks are different from the human-computer dialog task proposed in~\cite{Jason:2016：NIPS,Jason:2016:CoRR_2,OriolVinyals:2015:CoRR}.
\section{Gated Recurrent Unit Networks}

Gated Recurrent Unit (GRU)~\citep{Kyunghyun:2014:EMNLP} is the basic building block of our model for IQA. GRU has been widely adopted for many NLP tasks, such as machine translation~\citep{Dzmitry:2015:ICLR} and language modeling~\citep{Wojciech:CoRR:2014}. GRU improves computational efficiency over Long Short-term Memory (LSTM)~\citep{Sepp:1997:NC} by removing the cell component and making each hidden state adaptively capture the dependencies over different time steps using reset and update gates. For each time step $t$ with input $\mathbf{x}^t$ and previous hidden state $\mathbf{h}^{t-1}$, we compute the updated hidden state $\mathbf{h}^t = GRU(\mathbf{h}^{t-1}, \mathbf{x}^t)$ by,
\begin{align*}
	\mathbf{r}^t = \sigma(\mathbf{U}_r\mathbf{x}^t + \mathbf{W}_r\mathbf{h}^{t-1} + \mathbf{b}_r), \\
	\mathbf{z}^t = \sigma(\mathbf{U}_z\mathbf{x}^t + \mathbf{W}_z\mathbf{h}^{t-1} + \mathbf{b}_z), \\
	\mathbf{\widetilde{h}}^t = tanh(\mathbf{U}_h\mathbf{x}^t + \mathbf{W}_h(\mathbf{r}^t \odot \mathbf{h}^{t-1}) + \mathbf{b}_h), \\
	\mathbf{h}^t = \mathbf{z}^t \odot \mathbf{h}^{t-1} +  (\mathbf{1}-\mathbf{z}^{t}) \odot \mathbf{\widetilde{h}}^t,
\end{align*}
\noindent where $\sigma$ is the sigmoid activation function, $\odot$ is an element-wise product, $\mathbf{U}_r, \mathbf{U}_z, \mathbf{U}_h \in \mathbb{R}^{K \times D}$, $\mathbf{W}_r, \mathbf{W}_z, \mathbf{W}_h \in \mathbb{R}^{K \times K}$, $\mathbf{b}_r, \mathbf{b}_z, \mathbf{b}_h \in  \mathbb{R}^{K \times 1}$, $K$ is the  hidden size and $D$ is the input dimension size.

\begin{figure*}[ht]
	\centering
	\includegraphics[width=0.82\textwidth]{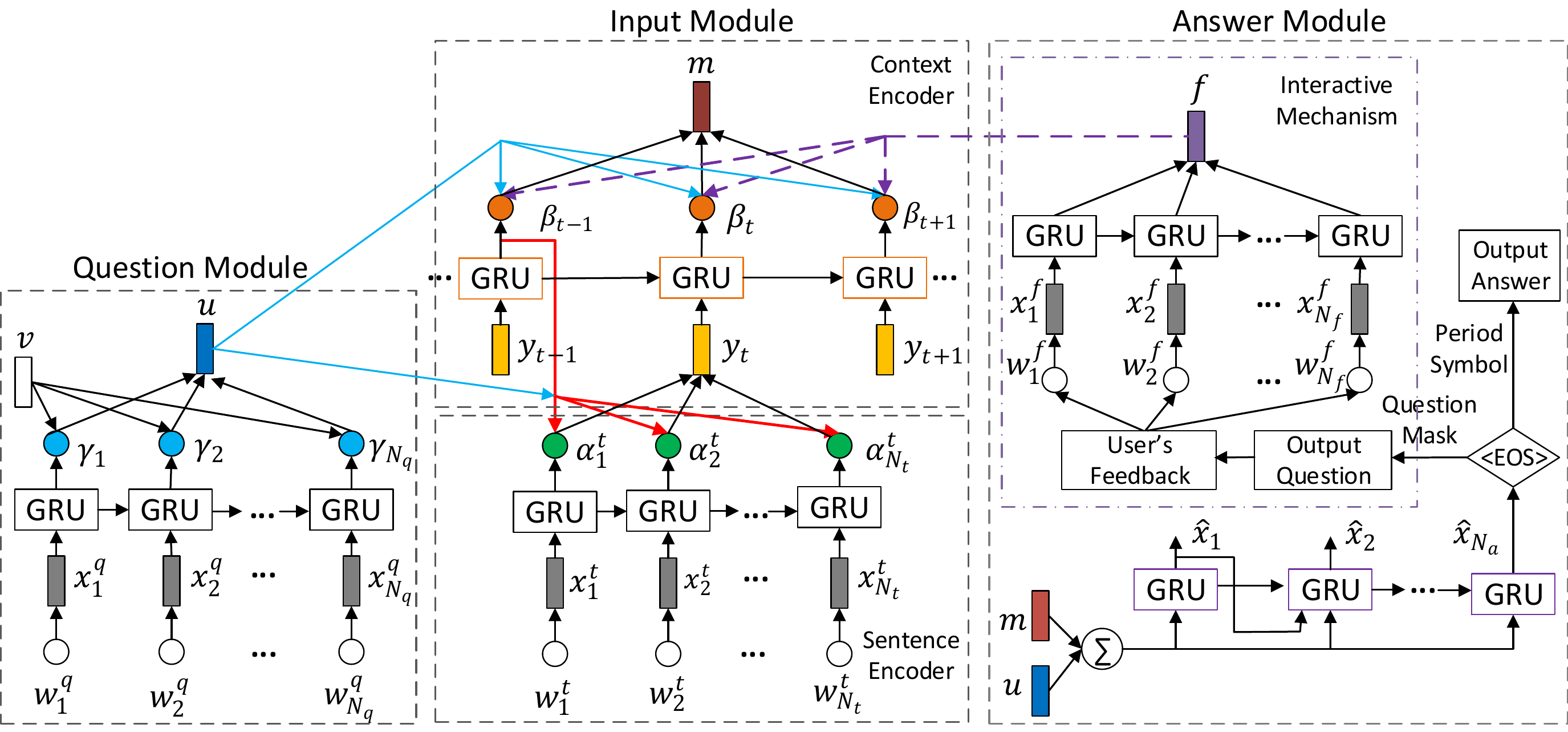}
	\vspace{-0.3cm}
	\caption{The illustration of the proposed model, consisting of a question module, an input module and an answer module.}
	\label{Fig:Model}
	\vspace{-0.4cm}
\end{figure*}

\section{Context-aware Attention Network}

In this section, we first illustrate the framework of our model CAN (Section~\ref{sec:framework}), including a question module (Section~\ref{sec:question_module}), an input module (Section~\ref{sec:input_module}),  and an answer module (Section~\ref{sec:answer_module}). We then describe each of these modules in detail. Finally, we elaborate the training procedure of CAN (Section~\ref{sec:training_proc}).

\subsection{Framework}\label{sec:framework}

\textbf{Problem Statement and Notation.} Given a story represented by $N$ {\bf input} sentences (or statements), i.e., $(l_1, \cdots, l_N)$, and a {\bf question} $q$, our goal is to generate an {\bf answer} $a$. Each sentence $l_t$ includes a sequence of $N_t$ words, denoted as $(w^t_1, \cdots, w^t_{N_t})$, and a question with $N_q$ words is represented as $(w^q_1, \cdots, w^q_{N_q})$. Let $V$ denote the size of vocabulary, including the words from each  $l_t$, $q$ and $a$, and end-of-sentence (EOS) symbols. In this paper, scalars, vectors and matrices are denoted by lower-case letters, boldface lower-case letters and boldface capital letters, respectively.

The whole framework of our model is shown in Figure \ref{Fig:Model}, consisting of the following three key parts:

\begin{itemize}[leftmargin=*]
\item \textbf{Question} Module: The question module encodes a target question into a vector representation.

\item \textbf{Input} Module: The input module encodes a set of input sentences into a vector representation.

\item \textbf{Answer} Module: The answer module generates an answer based on the outputs of question and input modules. Unlike conventional QA models, it has two choices, either to output an answer immediately or to interact with the user for further information. Hence, if the model lacks sufficient evidence for answer prediction based on existing knowledge, an interactive mechanism is enabled. Specifically, the model generates a supplementary question, and the user needs to provide a feedback, which is utilized to estimate an answer.
\end{itemize}

\subsection{Question Module}\label{sec:question_module}
Suppose a question is a sequence of $N_q$ words, we encode each word $w_j$ as a $K_w$-dimensional vector $\mathbf{x}_j^q$ using a learned embedding matrix $\mathbf{W}_w \in \mathbb{R}^{K_w \times V}$, i.e., $\mathbf{x}_j^q = \mathbf{W}_w[w_j]$, where $[w_j]$ is a one-hot vector associated with word $w_j$. The word sequence within a sentence significantly affects each word's semantic meaning due to its dependence on previous words. Thus, a GRU is employed by taking each word vector $\mathbf{x}_j^q$ as input and updating the corresponding hidden state $\mathbf{g}^q_j\in\mathbb{R}^{K_h \times 1}$ as follows:
\begin{align}
\mathbf{g}^q_j = GRU_w(\mathbf{g}^q_{j-1}, \mathbf{x}^q_j),
\end{align}
\noindent where the subscript of GRU is used to distinguish from other GRUs used in the following sections. The hidden state $\mathbf{g}^q_j$ can be regarded as the annotation vector of word $w_j$ by incorporating the word order information. We also explored a variety of encoding schema, such as LSTM and traditional Recurrent Neural Networks (RNN). However, LSTM is prone to over-fitting due to a large number of parameters, and traditional RNN has a poor performance because of exploding and vanishing gradients~\citep{Bengio:1994:TNN}.

In addition, each word contributes differently to the representation of a question. For example, in a question `Where is the football?', `where' and `football' play a critical role in summarizing this sentence. Therefore, an attention mechanism is introduced to generate a question representation by focusing on important words with informative semantic meanings. A positive weight $\gamma_j$ is placed on each word to indicate the relative importance of contribution to the question representation. Specifically, this weight is measured as the similarity of corresponding word annotation vector $\mathbf{g}_j^q$ and a word-level latent vector $\mathbf{v}\in\mathbb{R}^{K_h \times 1}$ for questions which is jointly learned during the training process. The question representation $\mathbf{u} \in \mathbb{R}^{K_c \times 1}$ is then generated by a sum of the word annotation vectors weighted by their corresponding importance weights, where we also use a linear projection to transform the aggregated representation vector from a sentence-level space to a context-level space as follows:
\begin{align}
	\gamma_j = softmax(\mathbf{v}^T\mathbf{g}_j^q),
	\label{Eq:QE_Attention_Weight} \\
	\mathbf{u} = \mathbf{W}_{ch}\sum_{j=1}^{N_q}\gamma_j\mathbf{g}_j^q + \mathbf{b}^{(q)}_{c},
\end{align}
\noindent where $softmax$ is taken to normalize the weights and  defined as $softmax(x_i)=\frac{\exp(x_i)}{\sum_{j'}\exp(x_{j'})}$, $\mathbf{W}_{ch} \in \mathbb{R}^{K_c \times K_h}$, and $\mathbf{b}_c^{(q)} \in \mathbb{R}^{K_c \times 1}$.

\subsection{Input Module}\label{sec:input_module}
Input module aims at generating representations for input sentences, including a sentence encoder and a context encoder. Sentence encoder computes the representation of a single sentence, and context encoder calculates an aggregated representation of a sequence of input sentences.

\subsubsection{\textbf{Sentence Encoder}}

For each input sentence $l_t$, containing a sequence of $N_t$ words $(w_1, \cdots, w_{N_t})$, similar to the question module, each word $w_i$ is embedded into a word space $\mathbf{x}_i^t \in \mathbb{R}^{K_w \times 1}$ through the shared learned embedding matrix $\mathbf{W}_w$, and a recurrent neural network is used to capture the context information from the words in the same sentence. Let $\mathbf{h}^t_i \in \mathbb{R}^{K_h \times 1}$ denote the hidden state which can be interpreted as the word annotation in the input space. A GRU computes each word annotation by taking the embedding vector as input and relying on previous hidden state,
\begin{align}\label{equ:1}
\mathbf{h}^t_i = GRU_w(\mathbf{h}^t_{i-1}, \mathbf{x}^t_i).
\end{align}
In Eq.~\ref{equ:1}, each word annotation vector takes its word order into consideration to learn its semantic meaning based on previous information within the current sentence through a recurrent neural network. A QA system is usually given multiple input sentences which often form a story together. A single word has different meaning in different stories. Learning a single sentence context at which a word is located is insufficient to understand the meaning of this word, especially when the sentence is placed in a story context. In other words, only modeling a sequence of words prior to the current word within the current sentence may lose some important information and result in the generation of inaccurate sentence representation. Hence, we take the whole context into account as well to appropriately characterize each word and well understand the current sentence's meaning. Suppose $\mathbf{s}_{t-1} \in \mathbb{R}^{K_c \times 1}$ is the annotation vector of previous sentence $l_{t-1}$, which will be introduced in the next section. To incorporate context information generated by previous sentences, we feed word annotation vector $\mathbf{h}^t_i$ and previous sentence annotation vector $\mathbf{s}_{t-1}$ into a two-layer MLP, through which a context-aware word vector $\mathbf{e}^t_i \in \mathbb{R}^{K_c \times 1}$ is obtained as follows:
\begin{align}
\mathbf{e}^t_i = \sigma(\mathbf{W}_{ee}tanh(\mathbf{W}_{es}\mathbf{s}_{t-1} + \mathbf{W}_{eh}\mathbf{h}_i^t + \mathbf{b}_e^{(1)}) + \mathbf{b}_e^{(2)}),
\label{Eq:WE_E}
\end{align}
\noindent where $\mathbf{W}_{ee}, \mathbf{W}_{es} \in \mathbb{R}^{K_c \times K_c}$ and $\mathbf{W}_{eh} \in \mathbb{R}^{K_c \times K_h}$ are weight matrices, and $\mathbf{b}_e^{(1)}, \mathbf{b}_e^{(2)} \in \mathbb{R}^{K_c \times 1}$ are the bias terms. It is worth noting that $\mathbf{s}_{t-1}$ is dependent on its previous sentence. Recursively, this sentence relies on its previous one as well. Hence, our model is able to encode the previous context. In addition, the sentence representation should emphasize those words which are able to address the question. Inspired by this intuition, another word level attention mechanism is introduced to attend informative words about the question for generating a sentence's representation. As the question representation is utilized to guide the word attention, a positive weight $\alpha^t_i$ associated with each word is computed as the similarity of the question vector $\mathbf{u}$ and the corresponding context-aware word vector $\mathbf{e}_i^t$.  Then the sentence representation $\mathbf{y}_t \in \mathbb{R}^{K_h \times 1}$ is generated by aggregating the word annotation vectors with different weights, and shown as follows,
\begin{align}
\alpha^t_i = softmax(\mathbf{u}^T\mathbf{e}_i^t), \\
\mathbf{y}_t = \sum_{i=1}^{N_t}\alpha^t_i\mathbf{h}_i^t.
\end{align}

\subsubsection{\textbf{Context Encoder}}
Suppose a story is comprised of a sequence of sentences, i.e., $(l_1,\cdots,l_N)$, each of which is encoded as a $K_h$-dimensional vector $\mathbf{y}_t$ through a sentence encoder. As input sentences have a sequence order, simply using their sentence vectors for context generation cannot effectively capture the entire context of the sequence of sentences. To address this issue, a sentence annotation vector is introduced to capture the previous context and this sentence's own meaning using a GRU. Given the sentence vector $\mathbf{y}_t$ and the state $\mathbf{s}_{t-1}$ of previous sentence, we get annotation vector $\mathbf{s}_t \in \mathbb{R}^{K_c \times 1}$ as follows:
\begin{align}
\mathbf{s}_t = GRU_s(\mathbf{s}_{t-1}, \mathbf{y}_t).
\end{align}
A GRU can learn a sentence's meaning based on previous context information. However, just relying on GRU at sentence level using simple word embedding vectors makes it difficult to learn the precise semantic meaning of each word in the story. Hence, we introduce a context-aware attention mechanism shown in Eq.~\ref{Eq:WE_E} to properly encode each word for the generation of sentence representation, which guarantees that each word is reasoned under an appropriate context.

Once the sentence annotation vectors $(\mathbf{s}_1,\cdots, \mathbf{s}_N)$ are obtained as described above, a sentence level attention mechanism is enabled to emphasize those sentences that are highly relevant to the question. We estimate each attention weight $\beta_t$ by the similarity between the question representation vector $\mathbf{u}$ and the corresponding sentence annotation vector $\mathbf{s}_t$ . Hence, the overall context representation vector $\mathbf{m}$ is  calculated by summing over all sentence annotation vectors weighted by their corresponding attention weights as follows,
\begin{align}
\beta_t = softmax(\mathbf{u}^T\mathbf{s}_t),\\
\mathbf{m} = \sum_{t=1}^{N}\beta_t\mathbf{s}_t.\label{Eq:CE_SR}
\end{align}
Similar to bidirectional RNN, our model can be extended to use another sentence-level GRU that moves backward through time beginning from the end of the sequence, but it does not have significant improvements in our experiments.

\subsection{Answer Module}\label{sec:answer_module}
The answer module utilizes a decoder to generate an answer, and has two output cases depending on both the question and the context. One case is to generate an answer immediately after receiving the context and question information. The other one is to generate a supplementary question and then uses the user's feedback to predict an answer. The second case requires an interactive mechanism.

\subsubsection{\textbf{Answer Generation}}
Given the question representation $\mathbf{u}$ and the context representation $\mathbf{m}$, another GRU is used as the decoder to generate a sentence as the answer. To use $\mathbf{u}$ and $\mathbf{m}$ together, we sum these vectors rather than concatenating them to reduce the total number of parameters. Suppose $\hat{\mathbf{x}}_{k-1} \in \mathbb{R}^{K_w \times 1}$ is the predicted word vector in last step, GRU updates the hidden state $\mathbf{z}_k \in \mathbb{R}^{K_o \times 1}$ as follows,
\begin{align}
\hat{\mathbf{x}}_{k} \overset{\mathbf{W}_w}{=} softmax(\mathbf{W}_{od}\mathbf{z}_k + \mathbf{b}_o), \\
\mathbf{z}_k = GRU_d(\mathbf{z}_{k-1}, [\mathbf{m}+\mathbf{u};\hat{\mathbf{x}}_{k-1}]),
\end{align}
\noindent where $\mathbf{W}_{od} \in \mathbb{R}^{V \times K_o}$, $\mathbf{b}_{o} \in \mathbb{R}^{V \times 1}$, $[\cdot;\cdot]$ indicates the concatenation operation of two vectors, and $\overset{\mathbf{W}_w}{=}$ denotes the predicted word vector through the embedding matrix $\mathbf{W}_w$. Note that we require that each sentence ends with a special EOS symbol, including question mask and period symbol, which enables the model to define a distribution over sentences of all possible lengths.

\textbf{Output Choices.} In practice, the system is not always able to answer a question immediately based on its current knowledge due to the lack of some crucial information bridging the gap between the question and the context knowledge, i.e., incomplete information. Therefore, we allow the decoder to make a binary choice, either to generate an answer immediately, or to enable an interactive mechanism. Specifically, if the model has sufficiently strong evidence for a successful answer prediction based on the well-learned context representation and question representation, the decoder will directly output the answer. Otherwise, the system generates a supplementary question for the user, where an example is shown in Table~\ref{table:example_ip}. At this time, this user needs to offer a feedback which is then encoded to update the sentence-level attentions for answer generation. This procedure is our interactive mechanism.

\begin{table} [ht]
\centering
\begin{tabular}{l|l}
	\hline
	\multirow{3}{*}{Problem} & The master bedroom is east of the garden.\\
	& The guest bedroom is east of the office.\\
	& Target Question: What is the bedroom east of? \\
	\hline
	& System: Which bedroom, master one or guest one?  \\
	Interactive & \quad\quad\quad\; \emph{(Supplementary Question)} \\
	\multirow{1}{*}{Mechanism} & User: \quad Master bedroom \quad  \emph{(User's Feedback)}\\
	& System: Garden \quad \emph{(Predicted Answer)}\\
	\hline
\end{tabular}
\caption{An example of interactive mechanism. }
\label{table:example_ip}
\vspace{-0.6cm}
\end{table}

The sentence generated by the decoder ends with a special symbol, either a question mask or a period symbol. Hence, this special symbol is utilized to make a decision. In other words, if EOS symbol is a question mask, the generated sentence is regarded as a supplementary question and an interactive mechanism is enabled; otherwise the generated sentence is the estimated answer and the prediction task is done. In the next section, we will present the details of the interactive mechanism.

\subsubsection{\textbf{Interactive Mechanism}}
The interactive process is summarized as follows: 1) The decoder generates a supplementary question; 2) The user provides a feedback; 3) The feedback is used for answer prediction for the target question. Suppose the feedback contains a sequence of words, denoted as $(w^f_1, \cdots, w^f_{N_f})$.  Similar to the input module, each word $w^f_d$ is embedded to a vector $\mathbf{x}^f_d$ through the shared embedding matrix $\mathbf{W}_w$. Then the corresponding annotation vector $\mathbf{g}^f_d \in \mathbb{R}^{K_h \times 1}$ is computed via a GRU by taking the embedding vector as input, and shown as follows:
\begin{align}
\mathbf{g}^f_d = GRU_w(\mathbf{g}^f_{d-1}, \mathbf{x}^f_d).
\end{align}
Based on the annotation vectors, a representation $\mathbf{f} \in \mathbb{R}^{K_h \times 1}$ can be obtained by a simple attention mechanism where each word is considered to contribute equally, and given by:
\begin{align}
\mathbf{f} = \frac{1}{N_f}\sum_{d=1}^{N_f}\mathbf{g}^f_d.
\end{align}
Our goal is to utilize the feedback representation $\mathbf{f}$ to generate an answer for the target question. The provided feedback improves the ability to answer the question by distinguishing the relevance of each input sentence to the question. In other words, the similarity of specific input sentences in the provided feedback make these sentences more likely to address the question. Hence, we refine the attention weight of each sentence shown in Eq.~\ref{Eq:CE_SR} after receiving the user's feedback, given by,
\begin{align}
\mathbf{r} = tanh(\mathbf{W}_{rf}\mathbf{f} + \mathbf{b}_r^{(f)}), \label{Eq:IM_MLP}\\
\beta_t = softmax(\mathbf{u}^T\mathbf{s}_t + \mathbf{r}^T\mathbf{s}_t)
\end{align}

\noindent where $\mathbf{W}_{rf} \in \mathbb{R}^{K_c \times K_h}$ and $\mathbf{b}_r^{(f)} \in \mathbb{R}^{K_c \times 1}$ are the weight matrix and bias vector, respectively.  Eq.~\ref{Eq:IM_MLP} is a one-layer neural network to transform the feedback representation to the context space. After obtaining the newly learned attention weights, we update the context representation using the soft-attention operation shown in Eq.~\ref{Eq:CE_SR}. This updated context representation and question representation will be used as the input for the decoder to generate an answer. Note that for simplifying the problem, we allow the decoder to only generate at most one supplementary question. In addition, one advantage of using the user's feedback to update the attention weights of input sentences is that we do not need to re-train the encoder once a feedback enters the system.

\subsection{Training Procedure}\label{sec:training_proc}
During training, all three modules share an embedding matrix. There are three different GRUs employed for sentence encoding, context encoding and answer/supplementary question decoding. In other words, the same GRU for sentence encoding is used to encode the question, input sentences and the user's feedback. The second GRU is applied to generate context representation and the third one is used as the decoder. Training is treated as a supervised sequence prediction problem by minimizing the cross-entropy between the answer sequence/the supplementary question sequence and the predictions.

\begin{table*} [t]
	\centering
	\begin{tabular}{| p{0.21\textwidth} | p{0.34\textwidth} | p{0.33\textwidth} |}
		\hline
		\textbf{IQA task 1:} & \textbf{IQA task 4:}  & \textbf{IQA task 7:} \\
		John journeyed to the garden. & The master bedroom is east of the garden.	&John grabbed the bread. \\
		Daniel moved to the kitchen. & The guest bedroom is east of the office.	& John grabbed the milk. \\
		& The guest bedroom is west of the hallway. & John grabbed the apple. \\
		& The bathroom is east of the master bedroom. & Sandra went to the bedroom. \\
		\ \ Q: Where is he? & \ \ Q: What is the bedroom east of? & \ \ Q: How many special objects is John holding? \\
		SQ: Who is he? & SQ: Which bedroom, master one or guest one? & SQ: What objects are you referring to? \\
		FB: Daniel    & FB: Master bedroom & FB: Milk, bread \\
		\ \ A: Kitchen  & \ \ A: Garden & \ \ A: Two \\
		\hline
	\end{tabular}
	\caption{Examples of three different tasks on the generated ibAbI datasets. ``Q'' indicates the target question. ``SQ'' is the supplementary question. ``FB'' refers to user's feedback. ``A'' is the answer.}
	\label{table:IQA_Samples}
	\vspace{-0.6cm}
\end{table*}

\section{Experiments}
In this section, we evaluate our approach with multiple datasets and make comparisons with state-of-the-art QA models.

\subsection{Experimental Setup}\label{Sec:Exp_Setup}
\textbf{Datasets.} In this paper, we use two types of datasets to evaluate the performance of our approach. One is a traditional QA dataset, where we use Facebook bAbI English 10k dataset~\citep{Jason:2015:CoRR} which is widely adopted in recent QA research~\cite{Caiming:2016:ICML,Ankit:2016:ICML,Sukhbaatar:2015:NIPS,Jason:2014:CoRR}. It contains 20 different types of tasks with emphasis on different forms of reasoning and induction. The second is our designed IQA dataset~\footnote{http://www.cs.toronto.edu/pub/cuty/IQAKDD2017}, where we extend bAbI by adding interactive QA and denote it as ibAbI. The reason for developing the ibAbI dataset is the absence of such IQA datasets with incomplete or ambiguous information in the QA research field. The settings of the ibAbI dataset follow the standard ones of bAbI datasets. Overall, we generate three ibAbI datasets based on task 1 (single supporting fact), task 4 (two argument relations), and task 7 (counting). The generated three ibAbI tasks simulate three different representative scenarios of incomplete or ambiguous information. Specifically, ibAbI task 1 focuses on ambiguous actor problem. ibAbI task 4 represents ambiguous object problem. ibAbI task 7 is to ask further information that assists answer prediction. Most of other IQA problems can be classified as one of these three tasks~\footnote{We do not need to modify each of the 20 bAbI task to make it interactive, because other extensions are either unnatural or redundant.}. Table \ref{table:IQA_Samples} shows three examples for our generated three ibAbI tasks, where the examples of supplementary question templates in different tasks are also provided.

To simulate real-world application scenarios, we mix IQA data and corresponding QA data together with different IQA ratios, where the IQA ratio is ranging from $0.3$ to $1$ (with step as $0.1$) and denoted as $R_{IQA}$. For example, in task $1$, we randomly pick $R_{IQA} \times 100$ percent data from ibAbI task 1, and then randomly select the remaining data from bAbI task 1.  $R_{IQA} = 1$ indicates that the whole dataset only consists of IQA problems; otherwise (i.e., ranging from $0.3$ to $0.9$) it consists of both types of QA problems. Overall, we have three tasks for the ibAbI dataset, and eight sub-datasets with different mixing ratios $R_{IQA}$ for each task. Therefore, we have 24 experiments in total for IQA. In addition, 10k examples are used as training and another 1k examples are used as testing.

\textbf{Experiment Settings.}
We train our models using the Adam optimizer~\citep{Diederik:2014:CoRR}. Xavier initialization is used for all parameters except for  word embeddings, which utilize random uniform initialization ranging from $-\sqrt{3}$ to $\sqrt{3}$. The learning rate is set as $0.001$. The grid search method is utilized to find optimal parameters, such as batch size and hidden dimension size and etc.

\subsection{Baseline Methods}
To demonstrate the effectiveness of our approach CAN, we compare it with the following four state-of-the-art models:

\begin{itemize}[leftmargin=*]
\item \textbf{DMN+:} It improves Dynamic Memory Networks~\citep{Ankit:2016:ICML} by using stronger input and memory modules~\cite{Caiming:2016:ICML}. 

\item \textbf{MemN2N:} This is an extension of Memory Network with weak supervision as proposed in~\cite{Sukhbaatar:2015:NIPS}. 

\item \textbf{EncDec:} We extend the encoder-decoder framework~\citep{Kyunghyun:2014:EMNLP} to solve QA tasks as a baseline method. EncDec uses the concatenation of statements and questions as input sentence to a GRU encoder, where the last hidden state is used as context representation, and employs another GRU as decoder.

\item \textbf{EncDec+IQA:} We extend EncDec to use our proposed interactive mechanism shown in Section~\ref{sec:answer_module} to evaluate the performance of our IQA concept in solving IQA problems. The difference is that after generating supplementary question, the  provided feedback by user is appended to the input sequence which is then encoded by the encoder again. The second output generated by the decoder is regarded as the prediction answer.
\end{itemize}
\noindent DMN+, MemN2N and EncDec are conventional QA models, while EncDec+IQA is purposely designed within our proposed IQA framework which can be viewed as an IQA base model.

\begin{table} [ht]
	\small
	\begin{center}
		\begin{tabular}{l c c c c}
			\hline
			Task & CAN+QA & DMN+ & MemN2N & EncDec\\
			\hline
			1 - Single Supporting Fact & \textbf{0.0} & \textbf{0.0} & \textbf{0.0} & 52.0\\
			2 - Two Supporting Facts & \textbf{0.1} & 0.3 & 0.3 & 66.1  \\
			3 - Three Supporting Facts & \textbf{0.2} & 1.1 & 2.1 & 71.9 \\
			4 - Two Arg. Relations & \textbf{0.0} & \textbf{0.0} & \textbf{0.0} & 29.2     \\
			5 - Three Arg. Relations & \textbf{0.4} & 0.5 & 0.8 & 14.3   \\
			6 - Yes/No Questions & \textbf{0.0} & \textbf{0.0} & 0.1 & 31.0       \\
			7 - Counting & \textbf{0.3} & 2.4 & 2.0 & 21.8 \\
			8 - Lists/Sets & \textbf{0.0} & \textbf{0.0} & 0.9 & 27.6 \\
			9 - Simple Negation & \textbf{0.0} & \textbf{0.0} & 0.3 & 36.4\\
			10 - Indefinite Knowledge & \textbf{0.0} & \textbf{0.0} & \textbf{0.0} & 36.4\\
			11 - Basic Coreference & \textbf{0.0} & \textbf{0.0} & 0.1 & 31.7 \\
			12 - Conjunction & \textbf{0.0} & \textbf{0.0} & \textbf{0.0} & 35.0 \\
			13 - Compound Coref. & \textbf{0.0} & \textbf{0.0} & \textbf{0.0} & 6.80 \\
			14 - Time Reasoning & \textbf{0.0} & 0.2 & 0.1 & 67.2\\
			15 - Basic Deduction & \textbf{0.0} & \textbf{0.0} & \textbf{0.0} & 62.2 \\
			16 - Basic Induction & \textbf{43.0} & 45.3 & 51.8 & 54.0 \\
			17 - Positional Reasoning & \textbf{0.2} & 4.2 & 18.6 & 43.1 \\
			18 - Size Reasoning & \textbf{0.5} & 2.1 & 5.3 & 6.60 \\
			19 - Path Finding & \textbf{0.0} & \textbf{0.0} & 2.3  & 89.6 \\
			20 - Agent’s Motivations & \textbf{0.0} & \textbf{0.0} & \textbf{0.0} & 2.30 \\
			\hline
			No. of failed tasks  & \textbf{1} & 5 & 6 & 20 \\
			\hline
		\end{tabular}
		\caption{Performance comparison of various models in terms of test error rate (\%) and the number of failed tasks on a conventional QA dataset. }
		\label{table:Exp_QA}
		\vspace{-0.9cm}
	\end{center}
\end{table}

\begin{table*} [ht]
	\centering
	\begin{tabular}{| p{0.3\textwidth} | p{0.05\textwidth} | p{0.04\textwidth} | p{0.3\textwidth} | p{0.05\textwidth} | p{0.04\textwidth} |}
		\hline
		Story & Support & Weight & Story & Support & Weight \\
		\hline
		Line 1: Mary journeyed to the office. & & 0.00 & Line 1: John went back to the  kitchen. & &\\
		$\cdots$ & & &  $\cdots$ & & \\
		$\cdots$ & & &  Line 13 : Sandra  grabbed the  apple there. & yes & \textbf{0.14}  \\
		Line 48: Sandra grabbed the apple there. & yes & \textbf{0.13} &  $\cdots$ & & \\
		Line 49: Sandra dropped the apple. & yes & \textbf{0.85} &   Line 29: Sandra  left the  apple. & yes & \textbf{0.79}\\
		Line 50: $\cdots$ & &  &  Line 30: $\cdots$ & & \\
		\hline
		\multicolumn{3}{|c|}{What is Sandra carrying? Answer: nothing Prediction: nothing} & \multicolumn{3}{c|}{What is Sandra carrying? Answer: nothing Prediction: nothing} \\
		\hline
	\end{tabular}
	\caption{Examples of our model's results on QA tasks. Supporting facts are shown, but our model does not use them during training. ``Weight'' indicates  attention weight of a sentence. Our model can locate correct supporting sentences in long stories.}
	\label{table:Exp_Long_Story}
	\vspace{-0.7cm}
\end{table*}

\begin{table} [ht]
	\centering
	\begin{tabular}{|l|}
		\hline
		The red square is below the triangle. \\
		The pink rectangle is to the left of the red square.\\
		Q: Is the triangle above the pink rectangle? \\
		A: yes \\
		\hline
		\hline
		The box is bigger than the suitcase. \\
		The suitcase fits inside the container.\\
		The box of chocolates fits inside the container.\\
		The container fits inside the chest. \\
		The chocolate fits inside the suitcase. \\
		Q: Is the chest bigger than the suitcase? \\
		A: yes \\
		\hline
	\end{tabular}
	\caption{Examples of bAbI task 17 (top) and 18 (bottom), where our model predicts correct answers while MemN2N makes wrong predictions.}
	\label{table:QA_Example_tasks}
	\vspace{-0.8cm}
\end{table}

\begin{table} [ht]
	\small
	\begin{center}
		\begin{tabular}{l| c c c c c}
			\hline
			Task & CAN+IQA &EncDec+IQA & DMN+ & MemN2N & EncDec\\
			\hline
			Task 1 & \textbf{0} & 6 & 8 & 8 & 8\\
			Task 4 & \textbf{0} & 8 & 8 & 8 & 8 \\
			Task 7 & \textbf{2} & 7 & 8 & 8 & 8\\
			\hline
		\end{tabular}
		\caption{Performance comparison of various models from the number of failed datasets for each task in the IQA setting. Each task has eight datasets with different $R_{IQA}$.}
		\label{Table:IQA_Failed_Datasets}
		\vspace{-0.8cm}
	\end{center}
\end{table}

\subsection{Performance of Question Answering}

In this section, we evaluate different models' performance for answer prediction based on the traditional QA dataset (i.e., bAbI-10k). For this task, our model (denoted as CAN+QA) does not use the interactive mechanism. As the output answers for this dataset only contain a single word, we adopt test error rate as evaluation metric. For DMN+ and MemN2N methods, we select the best performance over bAbI dataset reported in ~\citep{Caiming:2016:ICML}.
The results of various models
are reported in Table~\ref{table:Exp_QA}. We summarize the following observations:
\begin{itemize}[leftmargin=*]
	\item Our approach is better than all baseline methods on each individual task. For example, it reduces the error rate by $4\%$ compared to DMN+ in task 17, and compared to MemN2N, it reduces the error rate by, $18.4\%$ and $4.8\%$, respectively, on task 17 and 18. If using $1\%$ error rate as cutoff, our model only fails on $1$ task while DMN+ fails on $5$ tasks and MemN2N fails on $6$ tasks. Our model can achieve better performance mainly because our context-aware approach can model the semantic logic flow of statements. Table~\ref{table:QA_Example_tasks} shows two examples in task 17 and 18, where MemN2N predicts incorrectly while CAN+QA can make correct predictions. In these two examples, the semantic logic determines the relationship between two objects mentioned in the question, such as chest and suitcase. In addition, \cite{Ankit:2016:ICML} has shown that memory networks with multiple hops are better than the one with a single hop. However, our strong results demonstrate that our approach even without multiple hops has more accurate context modeling than previous models.

	\item EncDec performs the worst amongst all models over all tasks. EncDec concatenates the statements and questions as a single input, resulting in the difficulty of training the GRU. For example, EncDec performs terribly on task 2 and 3 because these two tasks have longer inputs than other tasks.
	
	\item The results of DMN+ and MemN2N are much better than EncDec. It is not surprising that they outperform EncDec, because they are specifically designed for QA and do not suffer from the problem mentioned above by treating input sentences separately.

	\item All models perform poorly on task 16. \citet{Caiming:2016:ICML} points out that MemN2N with a simple update for memory could achieve a near perfect error rate of $0.4$ while a more complex method will lead to a much worse result. This shows that a sophisticated modeling method makes it difficult to achieve a good performance in certain simple tasks with such limited data. This could be a possible reason explaining the poor performance of our model on this specific task as well.

\end{itemize}
In addition, different from MemN2N, we use a GRU to capture the semantic logic flow of input sentences, where the sentence-level attention on relevant sentences could be weakened by the influence of unrelated sentences in a long story. Table~\ref{table:Exp_Long_Story} shows two examples of our results with long stories. From the attention weights, we can see that our approach can correctly identify relevant sentences in long stories owing to our powerful context modeling.

\begin{figure*}[t]
	\centering
	\subfigure[IQA Task 1]{
		\includegraphics[width=0.3\textwidth]{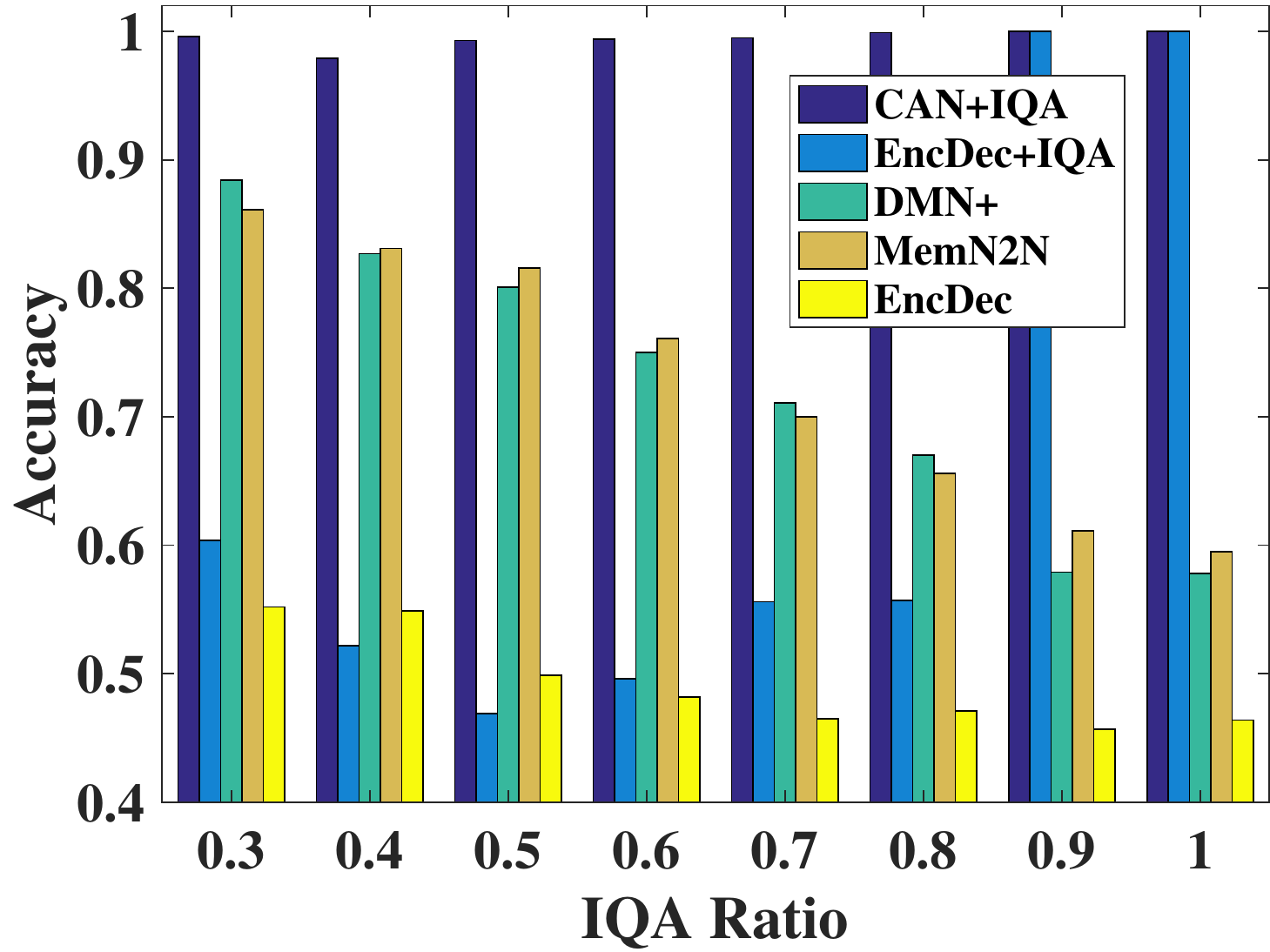}
	}
	\subfigure[IQA Task 4]{
		\includegraphics[width=0.3\textwidth]{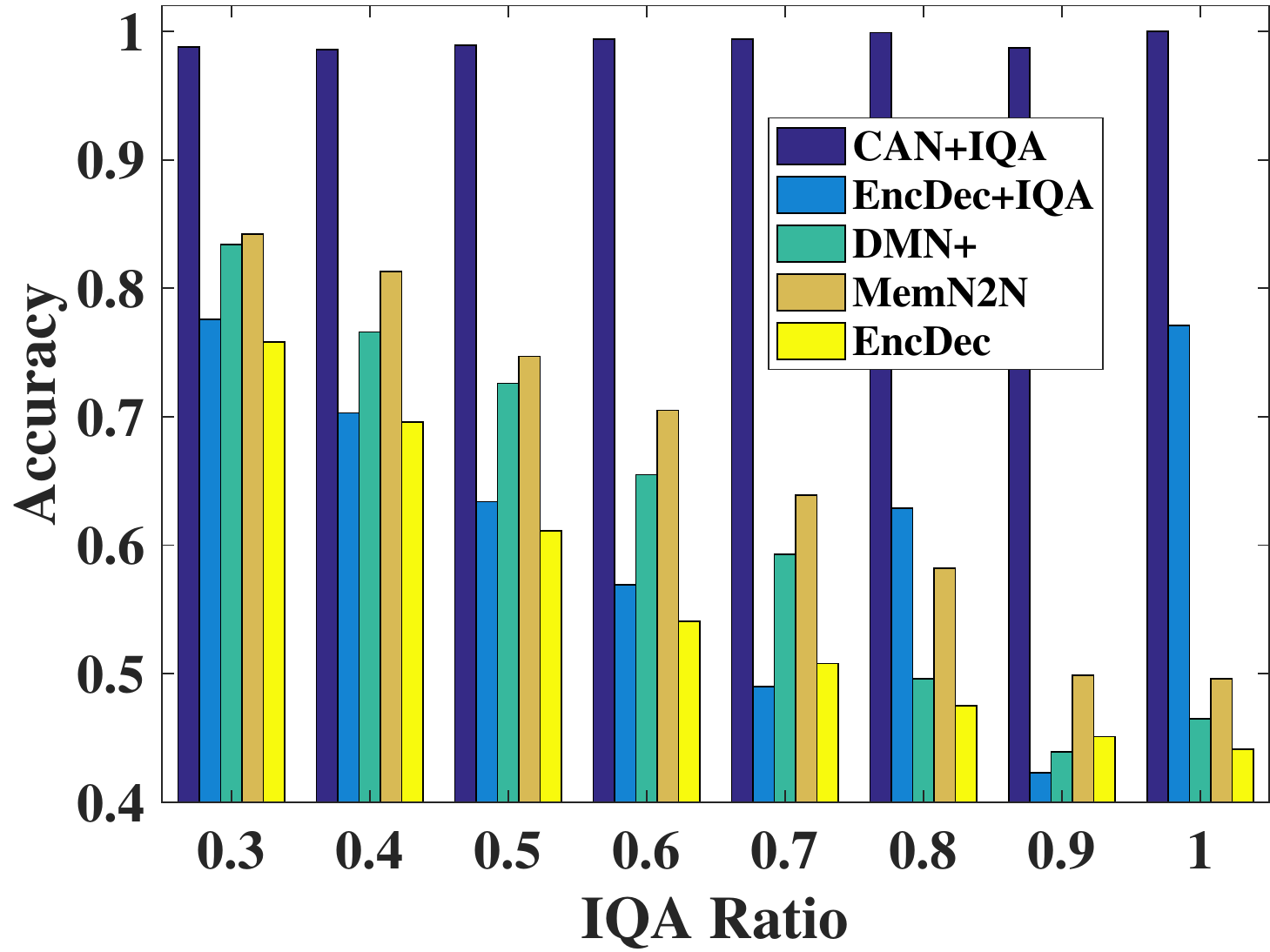}
	}
	\subfigure[IQA Task 7]{
		\includegraphics[width=0.3\textwidth]{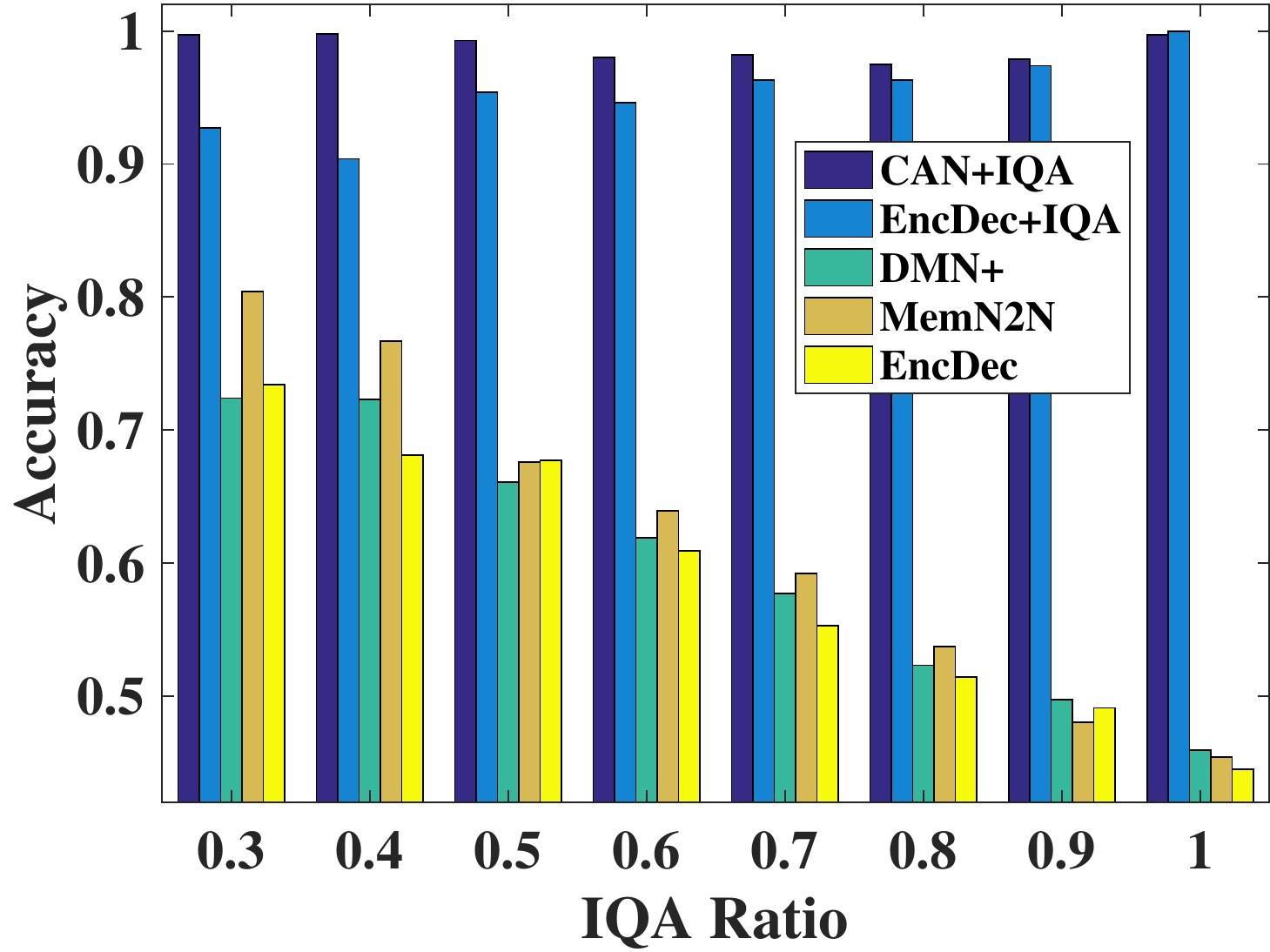}
	}
	\vspace{-0.4cm}
	\caption{Performance comparison of various models in terms of accuracy on IQA datasets with different IQA ratios.}
	\label{Fig:Exp_IQA}
	\vspace{-0.3cm}
\end{figure*}

\begin{table*} [ht]
	\small
	\begin{center}
		\begin{tabular}{p{0.3\textwidth} | c | p{0.2\textwidth}| p{0.1\textwidth} | p{0.1\textwidth}}
			\hline
			\multirow{2}{*}{Input Sentences} & \multirow{2}{*}{Support} & \multirow{2}{*}{QA Data} & \multicolumn{2}{c}{IQA Data} \\
			\cline{4-5}
			& & & Before IM & After IM \\
			\hline
			Mary journeyed to the kitchen.   &     & $0.00$ & \cellcolor[RGB]{1, 113, 193}{$0.99$} &  $0.00$\\
			Sandra journeyed to the kitchen. &     & $0.00$ & $0.00$ & $0.00$ \\
			Mary journeyed to the bedroom.   &     & $0.00$ & $0.00$ & $0.00$ \\
			Sandra moved to the bathroom.    &     & $0.00$ & $0.00$ & $0.00$ \\
			Sandra travelled to the office.  & yes & \cellcolor[RGB]{2, 113, 193}{$0.99$} & $0.00$ & \cellcolor[RGB]{4, 114, 193}{$0.99$} \\
			Mary journeyed to the garden.    &     & $0.00$ & $0.00$ & $0.00$ \\
			Daniel travelled to the bathroom.&     & $0.00$ & $0.00$ & $0.00$ \\
			Mary journeyed to the kitchen.   &     & $0.00$ & $0.00$ & $0.00$ \\
			John journeyed to the office.    &     & $0.00$ & $0.00$ & $0.00$ \\
			Mary moved to the bathroom.      &     & $0.00$ & $0.00$ & $0.00$ \\
			\hline
			\multicolumn{2}{c|}{} & Q: Where is Sandra? & \multicolumn{2}{p{0.2\textwidth}}{\ \ Q: Where is she?} \\
			\multicolumn{2}{c|}{} & A: Office & \multicolumn{2}{p{0.2\textwidth}}{SQ: Who is she?} \\
			\multicolumn{2}{c|}{} & & \multicolumn{2}{p{0.2\textwidth}}{FB: Sandra} \\
			\multicolumn{2}{c|}{} & & \multicolumn{2}{p{0.2\textwidth}}{\ \ A: Office} \\
			\hline
		\end{tabular}
		\caption{Examples of sentence attention weights obtained by our model in both QA and IQA data. ``Before IM" indicates the sentence attention weights over input sentences before the user provides a feedback. ``After IM" indicates the sentence attention weights updated by user's feedback. The attention weights with value as $0.00$ are very small. The results show that our approach can attend the key relevant sentences for both QA and IQA problems.}
		\label{table:Exmaple_Att}
		\vspace{-0.6cm}
	\end{center}
\end{table*}

\subsection{Performance of Interactive Question Answering}\label{sec:exp_iqa_performance}

In this section, we evaluate the performance of various models based on IQA datasets (as described in Section~\ref{Sec:Exp_Setup}). For testing, we simulate the interactive procedure by randomly providing a feedback according to the generated supplementary question as user's input, and then predicting an answer. For example, when asking ``who is he?'', we randomly select a male's name mentioned in the story as feedback. Conventional QA baseline methods, i.e., DMN+, MemN2N, and EncDec, do not have interactive part, so they cannot use feedback for answer prediction. Our approach (CAN+IQA) and EncDec+IQA adopt the proposed interactive mechanism to predict answer. We compare our approach with baseline methods in terms of accuracy shown in Figure~\ref{Fig:Exp_IQA}. Using $2\%$ error rate as cut off, the number of failed datasets for each task is also reported in Table~\ref{Table:IQA_Failed_Datasets}.  From the results, we can achieve the following conclusions:
\begin{itemize}[leftmargin=*]
	\item Our method outperforms all baseline methods and has significant improvements over conventional QA models. Specifically, we can nearly achieve $0\%$ test error rate with $R_{IQA} = 1.0$ ; while the best result of conventional QA methods can only get $40.5\%$ test error rate. CAN+IQA benefits from more accurate context modeling, which allows it to correctly understand when to output an answer or require additional information. For those QA problems with incomplete information, it is necessary to gather the additional information from users. Randomly guessing may harm model's performance, which makes conventional QA models difficult to converge. But our approach uses an interactive procedure to obtain user's feedback for assisting answer estimation.
	
	\item EncDec+IQA can achieve a relatively better result than conventional QA models in the datasets with high IQA ratios, especially in task 7. It happens due to our proposed interactive mechanism, where feedback helps to locate correct answers. However, it does not separate sentences, so the long inputs make its performance dramatically decreases as $R_{IQA}$ decreases. This explains its poor performance in most datasets with low IQA ratios, where there exists a large number of regular QA problems.
	
	\item For the conventional QA methods, DMN+ and MemN2N perform similarly and do better than EncDec. Their similar performance is due to the limitation that they could not learn the accurate meaning of statements and questions with limited resource and then have trouble in training the models. But they are superior over EncDec as they treat each input sentence separately instead of modeling very long inputs.
\end{itemize}

In addition, we also quantitatively evaluate the quality of supplementary question generated by our approach where the details can be found in Appendix~\ref{appendix:SQ_evaluation}.

\subsection{Qualitative Analysis of Interactive Mechanism}
In this section, we qualitatively show the attention weights over input sentences generated by our model on both QA and IQA data.  We train our model (CAN+IQA) on task 1 of ibAbI dataset with $Q_{IQA}=0.9$, and randomly select one IQA example from the testing data. Then we do the prediction on this IQA problem. In addition, we change this instance to a QA problem by replacing the question ``Where is she?'' with ``Where is Sandra?'', and then do the prediction as well. The prediction results on both QA and IQA problems are shown in Table~\ref{table:Exmaple_Att}. From the results, we observe the following: 1) The attention that uses user's feedback focuses on the key relevant sentence while the attention without feedback only focuses on an unrelated sentence. This happens because utilizing user's feedback allows the model to understand a question better and locate the relevant input sentences. This illustrates the effectiveness of an interactive mechanism on addressing questions that require additional information. 2) The attention on both two problems can finally focus on the relevant sentences, showing the usefulness of our model for solving different types of QA problems.
\section{Conclusion}

In this paper, we presented a self-adaptive context-aware question answering model, CAN,  which learns more accurate context-dependent representations of words, sentences, and stories. More importantly, our model is aware of {\it what it knows} and {\it what it does not know} within the context of a story, and takes an interactive mechanism to answer a question.
Our developed CAN model and generated new IQA datasets will open a new avenue to explore for researchers in the QA community. In the future, we plan to employ more powerful attention mechanisms with explicit unknown state modeling and multi-round feedback-guided fine-tuning to make the model fully self-aware, self-adaptive, and self-taught. We also plan to extend our framework to harder co-reference problems such as the Winograd Schema Challenge and interactive visual QA tasks with uncertainty modeling. 

\vspace{-0.2cm}
\section*{Acknowledgments}
This work is partially supported by the NIH (1R21AA023975-01) and NSFC (61602234, 61572032, 91646204, 61502077).

\vspace{-0.2cm}
\bibliographystyle{ACM-Reference-Format}
\bibliography{sigproc}

\vspace{-0.1cm}
\appendix
\section{Supplementary Question Analysis}\label{appendix:SQ_evaluation}

We quantitatively evaluate the quality of supplementary question generated by IQA models on IQA dataset, i.e., CAN+IQA and EncDec+IQA. To test model's performance, we define some following metrics. Suppose the number of problems is $N$, and the number of problems having supplementary question is $N_s$. Then $N_a = N - N_s$ is the number of remaining problems. Let $SQueAcc=\frac{\hat{N}_s}{N_s}$ is the fraction of IQA problems which can be correctly estimated, and $AnsAcc=\frac{\hat{N}_a}{N_a}$ is the fraction of remaining problems which can be correctly estimated as QA problem. Thus, ${SQueAnsAcc=}\frac{\hat{N}_s + \hat{N}_a}{N}$ is the overall accuracy. In addition, the widely used BLEU~\citep{Papineni:2002:ACL} and METEROR~\citep{Satanjeev:2005:ACLW} are also adopted to evaluate the quality of generated supplementary question. The results of CAN+IQA and EncDec+IQA are presented in Table~\ref{table:SQue_Quality}.

From the results, we can observe that 1) Two models can almost correctly determine whether it is time to output a question or not; 2) Two models are able to generate the correct supplementary questions whose contents exactly match with the ground truth. There is no surprise that EncDec+IQA also performs well in generating question, because it is specifically designed for handling IQA problems.  However, its ability to predict answer is not as good as CAN+IQA (See in Section~\ref{sec:exp_iqa_performance}) because it models very long inputs instead of carefully separating input sentences.

\begin{table} [ht]
	\scriptsize
	\begin{center}
		\begin{tabular}{c | c c }
			\hline
			& CAN+IQA & EncDec+IQA \\
			\hline
			SQueAcc & 100\% & 100\% \\
			AnsAcc  & 100\% & 100\% \\
			SQueAnsAcc & 100\% & 100\% \\
			\hline
			BLEU-1 & 100\% & 100\% \\
			BLEU-4 & 100\% & 100\% \\
			\hline
			METEOR & 100\% & 100\% \\
			\hline
		\end{tabular}
		\caption{Performance comparison of the generated supplementary question quality with $R_{IQA}$ as $0.8$ in task 1. Both two methods achieve $100\%$ under all metrics in all tasks with other different $R_{IQA}$ values.}
		\label{table:SQue_Quality}
	\end{center}
\end{table}

\end{document}